\documentclass[11pt,a4paper]{article}
\usepackage{breakurl}
\usepackage[hyperref]{acl2018}
\usepackage{times}
\usepackage{latexsym}
\usepackage[fleqn]{amsmath}
\usepackage{url}
\usepackage{booktabs}
\usepackage{multirow}
\usepackage{graphicx}
\usepackage{caption}
\usepackage{subcaption}
\usepackage{mathabx}
\usepackage{xparse}
\usepackage{xspace}
\usepackage{mathpartir}
\usepackage{tikz}
\usepackage{tikz-dependency}
\usepackage{etoolbox}
\usepackage{array,booktabs,ragged2e}
\usepackage{amsthm}
\usepackage{setspace}
\usepackage{bm}
\usepackage[T1]{fontenc}
\usepackage{xcolor}
\usepackage{tikz-dependency}
\usepackage{tikz}

\usepackage{url}

\aclfinalcopy %
\title{Global Transition-based Non-projective Dependency Parsing}

\author{
  Carlos G{\'o}mez-Rodr{\'i}guez \\
  Universidade da Coru{\~n}a\\
  {carlos.gomez@udc.es} \\\And
  Tianze Shi \\
  Cornell University\\
  {tianze@cs.cornell.edu} \\\And
  Lillian Lee \\
  Cornell University\\
  {llee@cs.cornell.edu} \\}

\date{}

\def\namecite{\newcite}

\mathchardef\mhyphen="2D

\newcommand{\shift}{\ensuremath{\mathsf{sh}}\xspace}

\newcommand{\reftab}[1]{Table~\ref{#1}}
\newcommand{\reffig}[1]{Fig.~\ref{#1}}
\newcommand{\refsec}[1]{\S\ref{#1}}

\newtheorem*{theorem*}{Theorem}

\newtheorem*{lemma*}{Lemma}

\newcolumntype{R}[1]{>{\RaggedLeft\arraybackslash}p{#1}}

\newcommand{\posscite}[1]{\citeauthor{#1}'s \citeyearpar{#1}}

\newcommand{\mharg}[1]{\ensuremath{\mathit{MH}_{#1}}\xspace}
\newcommand{\mhfour}{\mharg{4}}
\newcommand{\mhthree}{\mharg{3}}
\newcommand{\mhk}{\mharg{k}}

\newcommand{\cost}{\ensuremath{\mathit{cost}}\xspace}
\newcommand{\head}{\ensuremath{\mathit{head}}\xspace}
\newcommand{\tseq}{\ensuremath{\mathbf{t}}}
\newcommand{\tgold}{\ensuremath{\mathbf{t^{{\rm *}}}}}
\newcommand{\that}{\ensuremath{\mathbf{\hat{t}}}}

\newcommand{\bivec}[1]{\ensuremath{\mathbf{#1}}}
\newcommand{\svec}[1]{\ensuremath{\bivec{s}_{#1}}}
\newcommand{\bvec}[1]{\ensuremath{\bivec{b}_{#1}}}
\newcommand{\wvec}[1]{\ensuremath{\bivec{w}_{#1}}}

\newcommand{\stddev}[2]{\ensuremath{#1_{\color{darkgray}{\pm #2}}}}

\newcommand{\sZbZ}{\ensuremath{\{\svec{0}, \bvec{0}\}}\xspace}
\newcommand{\sObZ}{\ensuremath{\{\svec{1}, \svec{0}, \bvec{0}\}}\xspace}
\newcommand{\sTbZ}{\ensuremath{\{\svec{2}, \svec{1}, \svec{0}, \bvec{0}\}}\xspace}

\newcommand{\oneec}{1EC\xspace}
\newcommand{\rowbest}[1]{{\bf #1}}

\newcommand{\combine}{\textsc{Combine}\xspace}
\newcommand{\link}{\textsc{Link}\xspace}

\newcommand{\transname}[1]{\ensuremath{\mathsf{#1}}}

\newlength{\savedmathindent}

\definecolor{mygray}{gray}{0.38}

\begin{document}
\maketitle

\begin{abstract}
\citet*{exact-minfeats} obtained state-of-the-art results for English and Chinese dependency
parsing by combining dynamic-programming implementations of transition-based
dependency parsers with a minimal set of bidirectional LSTM features. However,
their results were limited to projective parsing. In this paper, we extend
their
approach to support non-projectivity by providing the first practical
implementation of the \mhfour algorithm, an $O(n^4)$ mildly non-projective
dynamic-programming parser with very high coverage on non-projective treebanks.
To make \mhfour compatible with minimal transition-based feature sets, we introduce
a transition-based interpretation of
it in which parser items are
mapped to sequences of transitions.
We thus obtain
the first implementation of
global decoding for non-projective transition-based parsing,
and demonstrate empirically that it is more effective
than its projective counterpart
in parsing
a number of highly non-projective languages.
\end{abstract}

\section{Introduction}
\label{sec:intro}

Transition-based dependency parsers are a popular approach to natural language
parsing, as they achieve good results in terms of accuracy and efficiency
\cite{ym03,nivre-scholz,zhang-nivre,chen-manning-14,dyer-stack-lstm,andor-tb,kiperwasser-lstm}. Until very recently,
practical implementations of transition-based parsing were limited to
approximate inference, mainly in the form of greedy search or beam search.
While cubic-time exact inference algorithms for several well-known projective
transition systems had been known since the work of
\newcite{huang-dp} and \newcite{kuhlmann-dp}, they
had been considered of theoretical interest only due to their
incompatibility with rich feature models:
incorporation of complex features resulted in jumps
in asymptotic runtime complexity
to impractical levels.

However, the recent popularization of
bi-directional long-short term memory networks
\cite[bi-LSTMs;][]{bilstm}
to derive feature
representations for parsing,
given
their capacity to capture long-range
information,
has demonstrated that one may not need to use complex feature models to obtain good
accuracy \cite{kiperwasser-lstm,cross-lstm}. In this context,
\newcite{exact-minfeats} presented an implementation of the exact inference
algorithms of \newcite{kuhlmann-dp} with a minimal set of only two
bi-LSTM-based feature
vectors.
This not only kept the complexity cubic, but also obtained
state-of-the-art results in English and Chinese parsing.

While
their approach provides both accurate parsing and the flexibility
to use any of greedy, beam, or exact decoding with the same underlying transition
systems, it does not support non-projectivity. Trees with crossing dependencies
make up a significant portion of many treebanks, going as high as 63\% for the
Ancient Greek treebank in the Universal Dependencies\footnote{\url{http://universaldependencies.org/}}
(UD)
dataset
version 2.0 and averaging
around 12\% over all languages in
UD 2.0.
In this
paper, we extend
\posscite{exact-minfeats} approach to mildly non-projective parsing in what, to our knowledge,
is the first implementation of exact decoding for a non-projective
transition-based parser.

As in the projective case, a mildly non-projective decoder has been
known for several years \cite{cohen-dp}, corresponding to a variant of the
transition-based parser of \citet{attardi}.
However, its $O(n^7)$ runtime --- or the $O(n^6)$ of a recently introduced improved-coverage
variant  \citep{attardi-dp-n6} --- is still prohibitively costly in practice.
Instead, we seek a more efficient algorithm to adapt, and
thus
develop a transition-based interpretation of
\posscite{gomez-nonproj-schemata}
$\mhfour$ dynamic
programming parser, %
which
has been shown to provide
very good
non-projective coverage in
$O(n^4)$ time \cite{coverage}.
 While the $\mhfour$ parser was originally
presented as a non-projective generalization of
the dynamic program
that later
led
to the arc-hybrid transition system
\cite{gomez-deductive,kuhlmann-dp}, its own relation to transition-based
parsing was not known. Here, we show that $\mhfour$ can be interpreted as
exploring a subset of the search space of a transition-based parser that
generalizes the arc-hybrid system, under a mapping that differs from the ``push
computation'' paradigm used by the previously-known dynamic-programming
decoders for transition systems. This allows us to extend
\newcite{exact-minfeats}'s work to non-projective parsing, by implementing
$\mhfour$ with a minimal set of transition-based features.

Experimental results show that our approach outperforms the projective approach
of \citet{exact-minfeats} and maximum-spanning-tree non-projective parsing on
the most highly non-projective languages in the CoNLL 2017
shared-task data that have a single treebank.
We also compare with the third-order 1-Endpoint-Crossing (\oneec) parser of
\citet{pitler-1ec-3o}, the only other practical implementation of
an exact mildly non-projective decoder
that we know of, which also runs in $O(n^4)$
but without a transition-based interpretation.
We obtain comparable results for these two algorithms,
in spite of
the fact that the $\mhfour$ algorithm is notably simpler than
\oneec.
The \mhfour parser remains effective in parsing projective treebanks,
while our baseline parser,
the fully non-projective maximum spanning tree algorithm,
falls behind due to
its unnecessarily large search space in parsing these languages.
Our code, including our re-implementation of the third-order \oneec parser
with neural scoring,
is available at \url{https://github.com/tzshi/mh4-parser-acl18}.

\section{Non-projective Dependency Parsing}
\label{sec:notation}

In dependency grammar, syntactic structures are modeled
as word-word asymmetrical subordinate relations among lexical entries \cite{kubler-book}.
These relations can be
represented in a graph.
For a sentence $w=w_1, ..., w_n$,
we first define a corresponding set of nodes
$\{0, 1, 2, ..., n\}$,
where $0$ is an artificial node denoting the root of the sentence.
Dependency relations are encoded by
edges
of the form
$(h, m)$,
where $h$ is the head and $m$ the modifier of the bilexical subordinate relation.\footnote{
To simplify exposition here, we only consider the unlabeled case.
We use a separately-trained labeling module
to obtain labeled parsing results in \refsec{sec:exp}.}

As is conventional, we assume two more properties
on dependency structures.
First, each word has exactly one syntactic head,
and second,
the structure is acyclic.
As a consequence, the edges
form a directed tree rooted at node $0$.

\begin{figure}[t]
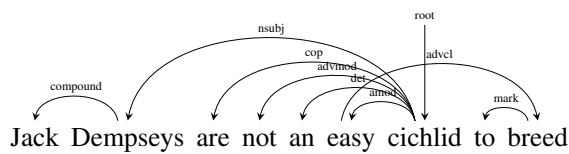

\centering

    \hspace*{-12pt}
    \begin{dependency}[theme=simple]
    \begin{deptext}
    Jack \& Dempseys \& are \& not \& an \& easy \& cichlid \& to \& breed \\
    \end{deptext}
    \depedge{2}{1}{compound}
    \depedge{7}{2}{nsubj}
    \depedge{7}{3}{cop}
    \depedge{7}{4}{advmod}
    \depedge{7}{5}{det}
    \depedge{7}{6}{amod}
    \deproot[edge unit distance=2.3ex]{7}{root}
    \depedge{9}{8}{mark}
    \depedge[edge start x offset=-8pt]{6}{9}{advcl}
    \end{dependency}

\caption{%
A non-projective dependency parse from the UD 2.0 English treebank.}
\label{fig:nonproj-ex}
\end{figure}

We say that a dependency structure is
{\em projective} if
it has no crossing edges.
While in the CoNLL and Stanford conversions of the English Penn Treebank, over  $99\%$ of the sentences are projective \cite{chen-manning-14}
--- see \reffig{fig:nonproj-ex} for
a non-projective English example --- for other languages' treebanks, non-projectivity is
a common occurrence (see Table \ref{tab:stats} for some statistics).  This paper
is targeted at learning parsers that can handle non-projective dependency trees.

\section{$\mhfour$ Deduction System and Its Underlying Transition System}
\label{sec:mh4}

\subsection{The $\mhfour$ Deduction System}

\begin{figure*}[h]

\[
\parbox{1.6cm}{\small{\textit{Axiom: }} $[0,1]$\\\large{\phantom{x}}}
\inferrule* [left=\text{Shift:}, right=$(h_m \le n)$]
{ [h_1,\ldots,h_m] }
{ [h_m,h_m+1] }
\hspace{0.3cm}
\renewcommand{\LeftTirName}[1]{\small{\textsc{#1}}}
\inferrule* [left=Combine:, right=$(p \le k)$]
{ [h_1,\ldots,h_m] \\ [h_m,h_{m+1},\ldots,h_p] }
{ [h_1,\ldots,h_p] }
\]
\vspace{-15pt}
\[
\hspace{-0.09cm}
\parbox{1.6cm}{\small{\textit{Goal: }} $[0,n+1]$\\\large{\phantom{x}}}
\inferrule* [left=Link:, right=${h_i} \rightarrow {h_j} ({1 \le i \le m} \wedge {1 < j < m} \wedge {j \neq i})$]
{ [h_1,\ldots,h_m] }
{ [h_1,\ldots,h_{j-1},h_{j+1},\ldots,h_m] }
\]

\caption{%
\mhk's deduction steps.}
\label{fig:steps-mhk}
\end{figure*}

The $\mhfour$ parser is the instantiation for $k = 4$ of
\posscite{gomez-nonproj-schemata}
more general $\mhk$ parser.
$\mhk$ stands for ``multi-headed with at most $k$ heads per item'':
items in its deduction system
take the form $[h_1, \ldots , h_p]$
for $p \le k$,
indicating
the existence of a forest of $p$ dependency subtrees headed
by $h_1, \ldots, h_p$
such that their yields are disjoint and
the union of their yields is the contiguous substring $h_1 \ldots h_p$ of the input. Deduction steps, shown in Figure \ref{fig:steps-mhk}, can be used to join two
such forests
that have an endpoint in common
via graph union
(\textsc{Combine}); or to add a dependency arc to a forest
that attaches an interior head as a dependent of any of the other heads (\textsc{Link}).

In the original formulation by \newcite{gomez-nonproj-schemata}, all valid items of the form $[i,i+1]$ are considered to be axioms.
In contrast, we follow \posscite{kuhlmann-dp} treatment of $\mhthree$:
we consider $[0,1]$ as the only axiom and include an extra \textsc{Shift} step to generate the rest of the items of that form.
Both formulations are equivalent, but including this \textsc{Shift} rule facilitates giving the parser a transition-based interpretation.

Higher values of $k$ provide wider coverage of non-projective structures at an asymptotic
runtime complexity of $O(n^k)$.
When $k$ is at its minimum value of 3, the parser covers exactly the set of projective trees, and in fact, it can be seen as a transformation\footnote{
Formally, it is a step refinement; see \citet{gomez-nonproj-schemata}.} of the deduction system described in \citet{gomez-deductive} that
gave rise to the projective arc-hybrid parser \cite{kuhlmann-dp}. For $k \ge 4$, the parser covers an increasingly larger set of non-projective structures. While a simple characterization of these sets has been lacking\footnote{
This is a common
issue with parsers based on the general idea of arcs between non-contiguous heads, such as those deriving from \citet{attardi}.}, empirical evaluation on a large number of treebanks \cite{coverage} has shown $\mhk$ to provide the best known tradeoff between asymptotic complexity and efficiency for $k>4$. When $k=4$, its coverage is second only to the 1-Endpoint-Crossing parser of \newcite{pitler-1ec}.
Both parsers fully cover well over 80\% of the non-projective trees observed in the studied treebanks.

\subsection{The $\mhfour$ Transition System}

\newcite{kuhlmann-dp} show how the items of a variant of $\mhthree$ can be given a transition-based interpretation
 under the ``push computation'' framework,
yielding
 the arc-hybrid projective transition system. However, such a derivation has not been made for the non-projective case ($k>3$), and the known techniques used to derive previous associations between tabular and transition-based parsers do not seem to be applicable in this case.
 The specific issue is that
 the deduction systems of \citet{kuhlmann-dp} and \citet{cohen-dp} have in common
 that the structure of their derivations is similar to that of a Dyck
(or balanced-brackets)
 language, where steps corresponding to shift transitions are balanced with those corresponding to reduce transitions. This makes it possible to group derivation subtrees, and the transition sequences that they yield, into
 ``push computations'' that increase the length of the stack by a constant amount. However, this does not seem possible in $\mhfour$.

Instead, we derive a transition-based interpretation of $\mhfour$ by a generalization of that of $\mhthree$ that departs from push computations.

To do so, we start with the $\mhthree$ interpretation of an item $[i,j]$
given by \newcite{kuhlmann-dp}. This item represents a set of computations (transition sequences) that start from a configuration of the form $(\sigma,i|\beta,A)$ (where $\sigma$ is the stack and
 $i|\beta$ is the buffer, with $i$ being the first buffer node) and take the parser to a configuration of the form
 $(\sigma|i,j|\beta',A)$. That is, the computation has the net effect of placing node $i$ on top of the previous contents of the stack, and it ends in a state where the first buffer element is $j$.

Under this item semantics, the \textsc{Combine} deduction step of the $\mhthree$ parser (i.e., the instantiation of the one in
\reffig{fig:steps-mhk} for $k=3$)
simply concatenates transition sequences.
The \textsc{Shift} step generates a sequence with a single arc-hybrid $\transname{sh}$ transition:
\[
\transname{sh}: (\sigma, h_m|\beta, A) \vdash (\sigma|h_m, \beta, A)
\]

\noindent
and the two possible instantiations of the \textsc{Combine} step when $k=3$ take the antecedent transition sequence and add a transition to it, namely, one of the two arc-hybrid reduce transitions. Written in the context of the node indexes used in Figure \ref{fig:steps-mhk}, these are the following:
\setlength{\savedmathindent}{\mathindent}
\setlength{\mathindent}{0cm}
\begin{multline*}
{\displaystyle (\sigma|h_1|h_2, h_3|\beta, A) \vdash (\sigma|h_1, h_3|\beta, A \cup \{h_3 \rightarrow h_2\})}\\
{\displaystyle (\sigma|h_1|h_2, h_3|\beta, A) \vdash (\sigma|h_1, h_3|\beta, A \cup \{h_1 \rightarrow h_2\})}
\end{multline*}
\normalsize
\noindent where
$h_1$ and $h_3$ respectively
can be simplified out to obtain the well-known arc-hybrid transitions:
\begin{multline*}
{\displaystyle \transname{la}: (\sigma|h_2, h_3|\beta, A) \vdash (\sigma, h_3|\beta, A \cup \{h_3 \rightarrow h_2\})}\\
{\displaystyle \transname{ra}: (\sigma|h_1|h_2, \beta, A) \vdash (\sigma|h_1, \beta, A \cup \{h_1 \rightarrow h_2\})}
\end{multline*}
Now, we assume the following generalization of the item semantics: an item $[h_1,\ldots,h_m]$ represents a set of computations that start from a configuration of the form $(\sigma,h_1|\beta,A)$ and lead to a configuration of the form
$(\sigma|h_1|\ldots|h_{m-1},h_m|\beta',A)$. Note that this generalization no longer follows the ``push computation'' paradigm of \citet{kuhlmann-dp} and \citet{cohen-dp}
because the number of nodes pushed
onto the stack depends on the value of $m$.

Under this item semantics, the \textsc{Shift} and \textsc{Combine} steps have the same interpretation as for $\mhthree$. In the case of the \textsc{Link} step, following the same reasoning as for the $\mhthree$ case, we obtain the following transitions:
\begin{multline*}
{\displaystyle \transname{la}: (\sigma|h_3, h_4|\beta, A) \vdash (\sigma, h_4|\beta, A \cup \{h_4 \rightarrow h_3\})}\\
{\displaystyle\transname{ra}: (\sigma|h_2|h_3, \beta, A) \vdash (\sigma|h_2, \beta, A \cup \{h_2 \rightarrow h_3\})}\\
{\displaystyle \transname{la'}: (\sigma|h_2|h_3, h_4|\beta, A) \vdash} \\
            {\displaystyle    \hspace*{1in} (\sigma|h_3, h_4|\beta, A \cup \{h_3 \rightarrow h_2\})}\\
{\displaystyle \transname{ra'}: (\sigma|h_1|h_2|h_3, \beta, A) \vdash} \\
            {\displaystyle \hspace*{1in}(\sigma|h_1|h_3, \beta, A \cup \{h_1 \rightarrow h_2\})}\\
{\displaystyle
\transname{la_2}:(\sigma|h_2|h_3, h_4|\beta, A) \vdash} \\
            {\displaystyle \hspace*{1in} (\sigma|h_3, h_4|\beta, A \cup \{h_4 \rightarrow h_2\})}\\
{\displaystyle \transname{ra_2}: (\sigma|h_1|h_2|h_3, \beta, A) \vdash} \\
             {\displaystyle \hspace*{1in} (\sigma|h_1|h_2, \beta, A \cup \{h_1 \rightarrow h_3\})}
\end{multline*}
\noindent These transitions
give us the $\mhfour$ transition system: a parser with four projective reduce transitions (\transname{la},\transname{ra},\transname{la'},\transname{ra'}) and two Attardi-like, non-adjacent-arc reduce transitions (\transname{la_2} and \transname{ra_2}).

It is worth mentioning that this $\mhfour$ transition system we have obtained is the same as one of the variants of Attardi's algorithm introduced by \namecite{attardi-dp-n6}, there called {\textsc{All${s_0s_1}$}\xspace}. However, in that paper they show that it can be tabularized in $O(n^6)$ using the push computation framework. Here, we have derived it as an interpretation of the $O(n^4)$ $\mhfour$ parser.

However, in this case the dynamic programming algorithm does not cover the full search space of the transition system:
while each item in the $\mhfour$ parser can be mapped into a computation of this $\mhfour$ transition-based parser,
the opposite is not true.
This tree:
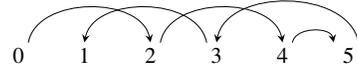
\begin{center}
{\small
\begin{dependency}[theme = simple]
\begin{deptext}[column sep=2em]
0 \& 1 \& 2 \& 3 \& 4 \& 5 \\
\end{deptext}
\depedge{1}{3}{}
\depedge{3}{5}{}
\depedge[edge end x offset=-6pt]{5}{6}{}
\depedge[edge start x offset=8pt]{6}{4}{}
\depedge{4}{2}{}
\end{dependency}
}
\end{center}
can be parsed by the transition system using the computation
\begin{multline*}
\transname{sh}(0); \transname{sh}(1); \transname{sh}(2); \transname{la_2}(3 \!\rightarrow\! 1); \transname{sh}(3); \transname{sh}(4);\\
\transname{la_2}(5 \!\rightarrow\! 3); \transname{sh}(5); \transname{ra}(4 \!\rightarrow\! 5); \transname{ra}(2 \!\rightarrow\! 4); \transname{ra}(0 \!\rightarrow\! 2)
\end{multline*}
\noindent but it is not covered by the dynamic programming algorithm, as no deduction sequence will yield an item representing this transition sequence. As we will see, this
issue
will not prevent us from implementing
a dynamic-programming parser  with transition-based scoring functions,
or from achieving good practical accuracy.

\setlength{\mathindent}{\savedmathindent}

\section{Model}
\label{sec:model}

Given the transition-based interpretation of the \mhfour\ system,
the learning objective becomes to
find a computation that gives the gold-standard parse.
For each sentence $w_1,\ldots,w_n$, we train
parsers to produce the transition sequence
$\tgold$ that corresponds to the annotated dependency structure.
Thus, the model consists of two components: a parameterized scorer $S(\tseq)$,
and a decoder that finds a sequence $\that$ as prediction based on the scoring.

As discussed by \citet{exact-minfeats},
there exists some tension between rich-feature scoring models and choices of decoders.
Ideally, a globally-optimal decoder finds the maximum-scoring transition sequence
$\that$
without
brute-force searching the exponentially-large output space.
To keep the runtime of
our exact decoder
at a practical low-order polynomial,
we
want its feature set to be minimal,
consulting as few
stack and buffer positions as possible.
In what follows, we use $s_0$ and $s_1$ to denote the top two stack items and $b_0$
and $b_1$ to denote the first two buffer items.

\subsection{Scoring and Minimal Features}

This section empirically explores the lower limit on the number of necessary positional features.
We experiment with both {\em local} and {\em global} decoding strategies.
The parsers take features extracted from
parser configuration $c$,
and score each valid transition $t$ with $S(t;c)$.
The {\em local} parsers greedily take transitions with the highest score until termination,
while the {\em global} parsers use the scores to find the globally-optimal solutions
$\that = \arg\max_{\tseq}{S(\tseq)}$,
where $S(\tseq)$ is the sum of scores for the component transitions.

Following prior work, we employ
bi-LSTMs
for compact feature representation.
A bi-LSTM runs in both directions on the input sentence, and assigns a context-sensitive vector encoding
to each token in the sentence: $\wvec{1},\ldots,\wvec{n}$.
When we need to extract features,
say, $\svec{0}$,
from a particular stack or buffer position, say $s_0$,
we directly use the bi-LSTM vector $\wvec{i_{s_0}}$, where $i_{s_0}$ gives the index of the subroot of $s_0$ into the sentence.
\citet{exact-minfeats} showed that feature vectors \sZbZ suffice for \mhthree.
\reftab{tab:minfeats-local}
and \reftab{tab:minfeats-global}
show the use of small feature sets for \mhfour,
 for {\em local} and {\em global} parsing models, respectively.
For a {\em local} parser to exhibit decent performance, we need at least \sObZ,
but adding $\svec{2}$ on top of that does not show any significant impact on the performance.
Interestingly, in the case of {\em global} models,
the two-vector feature set \sZbZ already suffices.
Adding \svec{1} to the global setting
(column ``Hybrid'' in \reftab{tab:minfeats-global})
seems attractive, but entails resolving a
technical challenge that we discuss in the following section.

\begin{table}[]
    \centering
    \begin{tabular}{rc@{\hspace{0.4em}}c@{\hspace{0.35em}}c@{\hspace{0.4em}}}
    \toprule
    Features & \sZbZ & \sObZ & \sTbZ \\
    \midrule
    UAS & $49.83$ & $85.17$ & $85.27$  \\
    \bottomrule
    \end{tabular}
    \caption{Performance of {\em local} parsing models with varying number of features.
    We report average UAS over $10$ languages on UD 2.0.
    }
    \label{tab:minfeats-local}
\end{table}

\subsection{Global Decoder}
\label{sec:decoding}

\begin{table}[]
    \centering
    \begin{tabular}{rcc}
    \toprule
    Features & $\{\svec{0}, \bvec{0}\}$ & Hybrid \\
    \midrule
    UAS & $86.79$ & $87.27$  \\
    \bottomrule
    \end{tabular}
    \caption{Performance of {\em global} parsing models with varying number of features.}
    \label{tab:minfeats-global}
\end{table}

In our transition-system interpretation of \mhk,
\shift transitions correspond to
\textsc{Shift}
and
reduce transitions reflect the \link steps.
Since the \textsc{Shift} conclusions
lose the contexts
needed to score the transitions,
we set the scores for all \textsc{Shift} rules to
zero and
delegate the scoring of the \shift transitions to the \combine steps,
as
as in \citet{exact-minfeats};
for example,
$$
\inferrule
{[h_1, h_2]: v_1\quad [h_2, h_3, h_4]: v_2}
{[h_1, h_2, h_3, h_4]: v_1 + v_2 + S(\shift; \{\bivec{h}_1, \bivec{h}_2\})}
$$
Here the transition sequence denoted by $[h_2, h_3, h_4]$ starts from a \shift,
with $h_1$ and $h_2$ taking the $s_0$ and $b_0$ positions.
If we further wish to access $s_1$, such information is not readily available in the deduction step,
apparently requiring extra bookkeeping that pushes the space and time complexity
to an impractical $O(n^4)$ and $O(n^5)$, respectively.
But, consider the scoring for the reduce transitions in the \link steps:
$$
\inferrule
{[h_1, h_2, h_3, h_4]: v}
{[h_1, h_2, h_4]: v + S(\transname{la}; \{\bivec{h}_2, \bivec{h}_3, \bivec{h}_4\})}
$$
$$
\inferrule
{[h_1, h_2, h_3]: v}
{[h_1, h_3]: v + S(\transname{la}; \{\bivec{h}_1, \bivec{h}_2, \bivec{h}_3\})}
$$
The deduction steps
\emph{already} keep indices for $s_1$
($h_2$ in the first rule, $h_1$ in the second)
and thus provide direct access without any modification.
To resolve the conflict between including $s_1$ for richer representations
and the unavailability of $s_1$ in scoring the \shift transitions in the \combine steps,
we propose
a hybrid scoring approach
--- we use features \sZbZ when scoring a \shift transition,
and features \sObZ for consideration of reduce transitions.
We call this method \mhfour-hybrid,
in contrast to \mhfour-two,
where we simply take \sZbZ for scoring all transitions.

\subsection{Large-Margin Training}
\label{sec:training}

We train the greedy parsers with
hinge loss, and
the global parsers with its structured version  \cite{taskar-margin}.
The loss function for each
sentence
is formally defined as:
$$\max_{\that}
\left(
{S(\that)+\cost(\tgold, \that)} - S(\tgold)
\right)
$$
\noindent where the margin $\cost(\tgold, \that)$ counts the number of mis-attached nodes
for taking sequence $\that$ instead of $\tgold$.
Minimizing this loss can be thought of as optimizing for the attachment scores.

The calculation of the above loss function can be solved as
efficiently as the deduction system if
the $\cost$ function decomposes into the dynamic program.
We achieve this by replacing the scoring of each reduce
step by its cost-augmented version:
$$
\inferrule
{[h_1, h_2, h_3, h_4]: v}
{[h_1, h_2, h_4]: v + S(\transname{la_2}; \{\bivec{h}_2, \bivec{h}_3, \bivec{h}_4\}) + \Delta}
$$
where $\Delta = \mathbf{1}(\head(w_{h_3}) \neq w_{h_4})$.
This loss function encourages the model to give higher contrast between gold-standard and wrong predictions,
yielding better generalization results.

\section{Experiments}
\label{sec:exp}

\paragraph{Data and Evaluation}

We experiment with the Universal Dependencies (UD) 2.0 dataset
used for
the
CoNLL 2017 shared task \cite{conll17}.
We restrict our choice of languages to be those with only
one training treebank,
for a better comparison with the shared task results.\footnote{
When multiple treebanks are available, one can develop domain transfer strategies,
which is not the focus of this work.
}
Among these languages, we pick the top 10 most non-projective languages.
Their basic statistics are listed in \reftab{tab:stats}.
For all
development-set results, we assume gold-standard tokenization and sentence delimitation.
When comparing to the shared task results on test sets,
we use the provided baseline UDPipe \cite{udpipe} segmentation.
Our models do not use part-of-speech tags or morphological tags as features,
but
rather leverage such information via stack propagation \cite{stack-prop},
i.e., we learn to predict them as a secondary training objective.
We report unlabeled attachment F1-scores (UAS) on the development sets for better focus on comparing our (unlabeled) parsing modules.
We report its labeled variant (LAS), the main metric of the shared task, on the test sets.
For each experiment setting, we ran the model with $5$ different random initializations,
and report the mean and standard deviation.
We detail the implementation details in the supplementary material.

\begin{table*}[htbp]
  \centering
    \begin{small}
    \begin{tabular}{rl|rr|ccc|ccc}
    \toprule
    \multirow{2}[0]{*}{Language} & \multirow{2}[0]{*}{Code} & \multirow{2}[0]{*}{\# Sent.} & \multirow{2}[0]{*}{\# Words} & \multicolumn{3}{c|}{Sentence Coverage ($\%$)} & \multicolumn{3}{c}{Edge Coverage ($\%$)} \\
      &   &   &   & Proj. $\downarrow$ & \mhfour & \oneec & Proj. & \mhfour & \oneec \\
    \midrule
        Basque & eu & $5{,}396$ & $72{,}974$ & $66.48$ & $91.48$ & $93.29$ & $95.98$ & $99.27$ & $99.42$ \\
        Urdu & ur & $4{,}043$ & $108{,}690$ & $76.97$ & $95.89$ & $95.77$ & $98.89$ & $99.83$ & $99.81$ \\
        Gothic & got & $3{,}387$ & $35{,}024$ & $78.42$ & $97.25$ & $97.58$ & $97.04$ & $99.73$ & $99.75$ \\
        Hungarian & hu & $910$ & $20{,}166$ & $79.01$ & $98.35$ & $97.69$ & $98.51$ & $99.92$ & $99.89$ \\
        Old Church Slavonic & cu & $4{,}123$ & $37{,}432$ & $80.16$ & $98.33$ & $98.74$ & $97.22$ & $99.80$ & $99.85$ \\
        Danish & da & $4{,}383$ & $80{,}378$ & $80.56$ & $97.70$ & $98.97$ & $98.60$ & $99.87$ & $99.94$ \\
        Greek & el & $1{,}662$ & $41{,}212$ & $85.98$ & $99.52$ & $99.40$ & $99.32$ & $99.98$ & $99.98$ \\
        Hindi & hi & $13{,}304$ & $281{,}057$ & $86.16$ & $98.38$ & $98.95$ & $99.26$ & $99.92$ & $99.94$ \\
        German & de & $14{,}118$ & $269{,}626$ & $87.07$ & $99.19$ & $99.27$ & $99.15$ & $99.95$ & $99.96$ \\
        Romanian & ro & $8{,}043$ & $185{,}113$ & $88.61$ & $99.42$ & $99.52$ & $99.42$ & $99.97$ & $99.98$ \\
    \bottomrule
    \end{tabular}%
\end{small}
    \caption{Statistics of selected training treebanks from Universal Dependencies 2.0 for the CoNLL 2017 shared task \cite{conll17},
    sorted by
    per-sentence projective ratio. }
    \label{tab:stats}%
\end{table*}%

\paragraph{Baseline Systems}

For comparison, we include three baseline systems with the same underlying feature representations and scoring paradigm.
All the following baseline systems are trained with the cost-augmented large-margin loss function.

The
{\em \mhthree parser} is the projective instantiation of the \mhk parser family.
This corresponds to the global version of
the
 arc-hybrid transition system \cite{kuhlmann-dp}.
We adopt
the minimal feature representation
\sZbZ,
following \citet{exact-minfeats}.
For this model, we also implement a greedy incremental version.

The
{\em edge-factored non-projective maximal spanning tree (MST) parser}
allows arbitrary non-projective structures.
This decoding approach has been shown to be very competitive in parsing non-projective treebanks \cite{mcdonald-mst},
and
was deployed in the
top-performing system at the CoNLL 2017 shared task \cite{dozat-conll}.
We score each edge individually,
with the features being the bi-LSTM vectors $\{\bivec{h}, \bivec{m}\}$,
where $h$ is the head, and $m$ the modifier of the edge.

The
{\em crossing-sensitive third-order \oneec parser}
provides a hybrid dynamic program for parsing 1-Endpoint-Crossing non-projective dependency trees
with higher-order factorization \cite{pitler-1ec-3o}.
Depending on whether
an edge is crossed,
we can access the modifier's grandparent $g$, head $h$, and sibling $si$.
We take their corresponding bi-LSTM features $\{\bivec{g}, \bivec{h}, \bivec{m}, \bivec{si}\}$ for scoring each edge.
This is a re-implementation of \citet{pitler-1ec-3o} with neural scoring functions.

\paragraph{Main Results}

\begin{table*}[htbp]
 \centering
    \begin{small}
    \begin{tabular}{r|ccccc|cc}
   \toprule
        & \multicolumn{5}{c|}{Global Models} & \multicolumn{2}{c}{Greedy Models}\\
    Lan. &  \mhthree  &  MST  &  \mhfour-two  &  \mhfour-hybrid  &  \oneec  &  \mhthree &  \mhfour \\
    \midrule
    eu & \stddev{82.07}{0.17} & \stddev{83.61}{0.16} & \stddev{82.94}{0.24} & \stddev{\rowbest{84.13}}{0.13} & \stddev{84.09}{0.19} & \stddev{81.27}{0.20} & \stddev{81.71}{0.33} \\
    ur & \stddev{86.89}{0.18} & \stddev{86.78}{0.13} & \stddev{86.84}{0.26} & \stddev{87.06}{0.24} & \stddev{\rowbest{87.11}}{0.11} & \stddev{86.40}{0.16} & \stddev{86.05}{0.18} \\
   got & \stddev{83.72}{0.19} & \stddev{84.74}{0.28} & \stddev{83.85}{0.19} & \stddev{84.59}{0.38} & \stddev{\rowbest{84.77}}{0.27} & \stddev{82.28}{0.18} & \stddev{81.40}{0.45} \\
    hu & \stddev{83.05}{0.17} & \stddev{82.81}{0.49} & \stddev{83.69}{0.20} & \stddev{\rowbest{84.59}}{0.50} & \stddev{83.48}{0.27} & \stddev{81.75}{0.47} & \stddev{80.75}{0.54} \\
    cu & \stddev{86.70}{0.30} & \stddev{88.02}{0.25} & \stddev{87.57}{0.14} & \stddev{88.09}{0.28} & \stddev{\rowbest{88.27}}{0.32} & \stddev{86.05}{0.23} & \stddev{86.01}{0.11} \\
    da & \stddev{85.09}{0.16} & \stddev{84.68}{0.36} & \stddev{85.45}{0.43} & \stddev{\rowbest{85.77}}{0.39} & \stddev{\rowbest{85.77}}{0.16} & \stddev{83.90}{0.24} & \stddev{83.59}{0.06} \\
    el & \stddev{87.82}{0.24} & \stddev{87.27}{0.22} & \stddev{87.77}{0.20} & \stddev{87.83}{0.36} & \stddev{\rowbest{87.95}}{0.23} & \stddev{87.14}{0.25} & \stddev{86.95}{0.25} \\
    hi & \stddev{93.75}{0.14} & \stddev{93.91}{0.26} & \stddev{93.99}{0.15} & \stddev{\rowbest{94.27}}{0.08} & \stddev{94.24}{0.04} & \stddev{93.44}{0.09} & \stddev{93.02}{0.10} \\
    de & \stddev{86.46}{0.13} & \stddev{86.34}{0.24} & \stddev{86.53}{0.22} & \stddev{86.89}{0.17} & \stddev{\rowbest{86.95}}{0.32} & \stddev{84.99}{0.26} & \stddev{85.27}{0.32} \\
    ro & \stddev{89.34}{0.27} & \stddev{88.79}{0.43} & \stddev{89.25}{0.15} & \stddev{\rowbest{89.53}}{0.20} & \stddev{89.52}{0.25} & \stddev{88.76}{0.30} & \stddev{87.97}{0.31} \\
    \midrule
    Avg. & $86.49$ & $86.69$ & $86.79$ & $\rowbest{87.27}$ & $87.21$ & $85.60$  & $85.27$ \\
    \bottomrule
    \end{tabular}%
    \end{small}
    \caption{Experiment results (UAS, $\%$) on the UD 2.0 development set.
    Bold: best result per language.}
    \label{tab:ud-dev}
\end{table*}

\reftab{tab:ud-dev} shows the development-set performance of our models as compared
with baseline systems.
MST considers non-projective structures, and thus
enjoys a
theoretical
advantage over projective \mhthree,
especially for the most non-projective languages.
However, it has a vastly larger output space,
making the selection of correct structures
difficult.
Further, the scoring is edge-factored, and does not take any structural contexts into consideration.
This tradeoff leads to the similar performance of MST comparing to \mhthree.
In comparison, both \oneec and \mhfour are mildly non-projective parsing algorithms,
limiting the size of the output space.
\oneec includes higher-order features that look at
tree-structural contexts;
\mhfour derives its features from parsing configurations of a transition system,
hence leveraging contexts within transition sequences.
These considerations explain their significant improvements over MST.
We also observe that \mhfour recovers more short dependencies than \oneec,
while \oneec is better at longer-distance ones.

In comparison to \mhfour-two,
the richer feature representation of \mhfour-hybrid helps in all
our languages.

Interestingly, \mhfour and \mhthree
react differently to switching from global to greedy models.
\mhfour covers more structures than \mhthree, and is naturally more capable in the global case,
even when the feature functions are the same (\mhfour-two).
However, its greedy
version is outperformed by \mhthree.
We
conjecture that this is because \mhfour explores only the same number of configurations as \mhthree,
despite the fact that introducing non-projectivity
expands the search space dramatically.

\paragraph{Comparison with CoNLL Shared Task Results
(\reftab{tab:ud-test})
}

\begin{table*}[!ht]
  \centering
  \begin{small}
  \begin{tabular}{r|ccc@{\hspace{0.4em}}c@{\hspace{0.8em}}c|>{\color{mygray}}c>{\color{mygray}}c>{\color{mygray}}c}
    \toprule
        & \multicolumn{5}{c|}{Same Model Architecture} & \multicolumn{3}{>{\color{mygray}}c}{For Reference} \\

    Lan. & \mhthree & MST & \mhfour-hybrid & & \oneec & Ensemble & CoNLL \#1 & CoNLL \#2 \\
    \midrule
    eu & \stddev{78.17}{0.33} & \stddev{79.90}{0.08} & \stddev{\rowbest{80.22}}{0.48} & $>$ & \stddev{80.17}{0.32} & $\rowbest{81.55}$ & $81.44$ & $79.61$ \\
    ur & \stddev{\rowbest{80.91}}{0.10} & \stddev{80.05}{0.13} & \stddev{80.69}{0.19} & $>$ & \stddev{80.59}{0.19} & $81.37$ & $\rowbest{82.28}$ & $81.06$ \\
   got & \stddev{67.10}{0.10} & \stddev{67.26}{0.45} & \stddev{\rowbest{67.92}}{0.29} & $>$ & \stddev{67.66}{0.20} & $\rowbest{69.83}$ & $66.82$ & $68.34$ \\
    hu & \stddev{76.09}{0.25} & \stddev{75.79}{0.36} & \stddev{\rowbest{76.90}}{0.31} & $>$ & \stddev{76.07}{0.20} & $\rowbest{79.35}$ & $77.56$ & $76.55$ \\
    cu & \stddev{71.28}{0.29} & \stddev{72.18}{0.20} & \stddev{72.51}{0.23} & $<$ & \stddev{\rowbest{72.53}}{0.27} & $\rowbest{74.38}$ & $71.84$ & $72.35$ \\
    da & \stddev{80.00}{0.15} & \stddev{79.69}{0.24} & \stddev{\rowbest{80.89}}{0.17} & $>$ & \stddev{80.83}{0.27} & $82.09$ & $\rowbest{82.97}$ & $81.55$ \\
    el & \stddev{85.89}{0.29} & \stddev{85.48}{0.25} & \stddev{\rowbest{86.28}}{0.44} & $>$ & \stddev{86.07}{0.37} & $87.06$ & $\rowbest{87.38}$ & $86.90$ \\
    hi & \stddev{89.88}{0.18} & \stddev{89.93}{0.12} & \stddev{90.22}{0.12} & $<$ & \stddev{\rowbest{90.28}}{0.21} & $90.78$ & $\rowbest{91.59}$ & $90.40$ \\
    de & \stddev{76.23}{0.21} & \stddev{75.99}{0.23} & \stddev{\rowbest{76.46}}{0.20} & $>$ & \stddev{76.42}{0.35} & $77.38$ & $\rowbest{80.71}$ & $77.17$ \\
    ro & \stddev{83.53}{0.35} & \stddev{82.73}{0.36} & \stddev{83.67}{0.21} & $<$ & \stddev{\rowbest{83.83}}{0.18} & $84.51$ & $\rowbest{85.92}$ & $84.40$ \\
    \midrule
    Avg. & $78.91$ & $78.90$ & $\rowbest{79.57}$ & $>$ & $79.44$ & $80.83$ & $\rowbest{80.85}$ & $79.83$ \\
    \bottomrule
    \end{tabular}%
    \end{small}
    \caption{Evaluation results (LAS, $\%$) on the test set using the CoNLL 2017 shared task setup.
    The best results for each language within each block are highlighted in bold.
    }
    \label{tab:ud-test}
\end{table*}%

We compare our models on the test sets, along with the best single model \cite[\#1;][]{dozat-conll} and the best ensemble model \cite[\#2;][]{shi-conll}
from the CoNLL 2017 shared task.
\mhfour outperforms \oneec in $7$ out of the $10$ languages.
Additionally, we take our non-projective parsing models (MST, \mhfour-hybrid, \oneec)
and combine them into an ensemble.
The average result is competitive with the best CoNLL submissions.
Interestingly, \citet{dozat-conll} uses fully non-projective parsing algorithms (MST),
and our ensemble system sees larger gains in the more non-projective languages,
confirming the potential benefit of global mildly non-projective parsing.

\begin{table*}[!ht]
  \centering
  \begin{small}
  \begin{tabular}{r|cccc|>{\color{mygray}}c>{\color{mygray}}c>{\color{mygray}}c}
    \toprule
        & \multicolumn{4}{c|}{Same Model Architecture} & \multicolumn{3}{>{\color{mygray}}c}{For Reference} \\
    Lan. & \mhthree & MST & \mhfour-hybrid & \oneec & Ensemble & CoNLL \#1 & CoNLL \#2 \\
    \midrule
    ja & \stddev{\rowbest{74.29}}{0.10} & \stddev{73.93}{0.16} & \stddev{74.23}{0.11} & \stddev{74.12}{0.12} & $74.51$ & $\rowbest{74.72}$ & $74.51$ \\
    zh & \stddev{\rowbest{63.54}}{0.13} & \stddev{62.71}{0.17} & \stddev{63.48}{0.33} & \stddev{\rowbest{63.54}}{0.26} & $64.65$ & $\rowbest{65.88}$ & $64.14$ \\
    pl & \stddev{86.49}{0.19} & \stddev{85.76}{0.31} & \stddev{\rowbest{86.60}}{0.26} & \stddev{86.36}{0.28} & $87.38$ & $\rowbest{90.32}$ & $87.15$ \\
    he & \stddev{61.47}{0.24} & \stddev{61.28}{0.24} & \stddev{\rowbest{61.93}}{0.22} & \stddev{61.75}{0.22} & $62.40$ & $\rowbest{63.94}$ & $62.33$ \\
    vi & \stddev{41.26}{0.39} & \stddev{41.04}{0.19} & \stddev{\rowbest{41.33}}{0.32} & \stddev{40.96}{0.36} & $\rowbest{42.95}$ & $42.13$ & $41.68$ \\
    bg & \stddev{87.50}{0.20} & \stddev{87.03}{0.17} & \stddev{\rowbest{87.63}}{0.17} & \stddev{87.56}{0.14} & $88.22$ & $\rowbest{89.81}$ & $88.39$ \\
    sk & \stddev{80.48}{0.22} & \stddev{80.25}{0.32} & \stddev{\rowbest{81.27}}{0.14} & \stddev{80.94}{0.25} & $82.38$ & $\rowbest{86.04}$ & $81.75$ \\
    it & \stddev{87.90}{0.07} & \stddev{87.26}{0.23} & \stddev{\rowbest{88.06}}{0.27} & \stddev{87.98}{0.19} & $88.74$ & $\rowbest{90.68}$ & $89.08$ \\
    id & \stddev{\rowbest{77.66}}{0.13} & \stddev{76.95}{0.32} & \stddev{77.64}{0.17} & \stddev{77.60}{0.18} & $78.27$ & $\rowbest{79.19}$ & $78.55$ \\
    lv & \stddev{69.62}{0.55} & \stddev{69.33}{0.51} & \stddev{\rowbest{70.54}}{0.51} & \stddev{69.52}{0.29} & $72.34$ & $\rowbest{74.01}$ & $71.35$ \\
    \midrule
    Avg. & $73.02$ & $72.55$ & $\rowbest{73.27}$ & $73.03$ & $74.18$ & $\rowbest{75.67}$ & $73.89$ \\
    \bottomrule
    \end{tabular}%
    \end{small}
    \caption{CoNLL 2017 test set results (LAS, $\%$) on the most projective languages
    (sorted by projective ratio; ja (Japanese) is fully projective).
    }
    \label{tab:ud-test-proj}
\end{table*}%

\paragraph{Results on Projective Languages (\reftab{tab:ud-test-proj})}
For completeness, we also test our models on the 10 most projective languages that have a single treebank.
\mhfour remains the most effective, but by a much smaller margin.
Interestingly, \mhthree, which is strictly projective, matches the performance of \oneec;
both outperform the fully non-projective MST by half a point.

\section{Related Work}
\label{sec:related}

Exact inference for dependency parsing can be achieved in cubic time if the model is restricted to
projective trees \cite{eisner96}. However, non-projectivity is needed for natural language parsers
to satisfactorily deal with linguistic phenomena like topicalization, scrambling and extraposition, which
cause crossing dependencies.
In UD 2.0,
68 out of 70 treebanks were reported
to contain non-projectivity \cite{wang17universal}.

However, exact inference has been shown to be intractable for models that support arbitrary non-projectivity,
except under strong independence assumptions \cite{mcdonald-complexity}. Thus, exact inference parsers that support
unrestricted non-projectivity are limited to edge-factored models \cite{mcdonald-mst,dozat-conll}.
Alternatives include
treebank transformation and pseudo-projective parsing \cite{projectivize,pseudo-proj},
approximate inference (e.g. \citet{mcdonald-pereira-06,attardi,nivre-swap,cov-nonmono})
or
focusing on sets of
dependency trees that allow only restricted
forms of non-projectivity.
A number of such sets, called mildly non-projective classes of trees, have been
identified
that
both exhibit good empirical coverage of the non-projective phenomena found in
natural
languages
and are known to have
polynomial-time exact parsing algorithms; see \citet{coverage} for a survey.

However, most of these algorithms have not been implemented in practice due to their prohibitive complexity.
For example, \citet{corro-wg1} report an implementation of the $\textit{WG}_1$ parser, a $O(n^7)$
mildly non-projective parser introduced in \citet{gomez-mildly-09}, but
it could not be run for real sentences of length greater than 20.
On the other hand, \newcite{gap-minding} provide an implementation of
an $O(n^5)$ parser for a mildly non-projective class of structures called gap-minding trees, but they need
to resort to aggressive pruning to make it practical, exploring only a part of the search space
in $O(n^4)$ time.
To the best of our knowledge, the only practical system that actually implements exact inference for
mildly non-projective parsing is the 1-Endpoint-Crossing
(1EC)
parser
of Pitler \citeyearpar{pitler-1ec,pitler-1ec-3o}, which runs
in $O(n^4)$ worst-case time like the $\mhfour$ algorithm used in this paper. Thus, the system presented here
is the second practical implementation of exact mildly non-projective parsing that has successfully been
executed on
real corpora.\footnote{
\citet{corro-wg1} describe a parser that enforces mildly non-projective constraints
(bounded block degree and well-nestedness), but it is an arc-factored model, so it is subject to the same strong
independence assumptions as maximum-spanning-tree parsers like \citet{mcdonald-mst} and does not support
the greater flexibility in scoring that is the main advantage of mildly non-projective parsers over these. Instead,
mild non-projectivity is exclusively used as a criterion to discard
nonconforming trees.
}

Comparing with \citet{pitler-1ec-3o}'s
1EC, our parser has the following disadvantages:
($-$1) It has slightly lower coverage, at least on the treebanks considered by \newcite{coverage}.
($-$2) The set of trees covered by $\mhfour$ has not been characterized with a non-operational definition, while
the set of 1-Endpoint-Crossing trees can be simply defined.

However, it also has the following advantages:
(+1) It can be
given a transition-based interpretation, allowing us to use transition-based scoring
functions and to implement the analogous algorithm with greedy or beam search apart from exact inference. No
transition-based interpretation is known for
1EC.
While a transition-based algorithm has been defined for a strict subset of 1-Endpoint-Crossing trees, called
2-Crossing Interval trees \cite{pitler-mcdonald}, this is a separate algorithm with no known mapping or relation
to
1EC or any other dynamic programming model.
Thus, we provide the first exact inference
algorithm for a non-projective transition-based parser with practical complexity.
(+2) It is conceptually much simpler, with one kind of item and two deduction steps, while the 1-Endpoint-Crossing
parser has five classes of items and several dozen distinct deduction steps. It is also a purely bottom-up parser,
whereas the 1-Endpoint-Crossing parser does not have the bottom-up property. This property is necessary for models that involve
compositional representations of subtrees \cite{dyer-stack-lstm}, and facilitates parallelization and partial parsing.
(+3) It can be easily generalized to $\mhk$ for $k>4$, providing higher coverage, with time complexity $O(n^k)$.
Out of the mildly non-projective parsers studied in \citet{coverage},
$\mhfour$ provides the maximum coverage with respect
to its complexity for $k>4$.
(+4)
As shown in \S\ref{sec:exp},
$\mhfour$
obtains slightly higher accuracy than 1EC
on average, albeit not by a conclusive margin.

It is worth noting that
1EC
has recently been extended to graph parsing by
\citet{kurtz-1ec}, \citet{kummerfeld-1ec}, and \citet{cao2017-1ec,cao2017-1ec-quasi}, with the latter providing a practical
implementation of a parser for 1-Endpoint-Crossing, pagenumber-2 graphs.

\section{Conclusion}
\label{sec:conclusion}

We have extended the
parsing architecture of \newcite{exact-minfeats} to non-projective
dependency parsing by implementing the \mhfour parser, a mildly non-projective $O(n^4)$ chart
parsing algorithm, using a minimal set of transition-based bi-LSTM features. For this purpose, we have established
a mapping between \mhfour items and transition sequences of an underlying non-projective transition-based parser.

To our knowledge, this is the first practical implementation
of exact inference for non-projective transition-based parsing. Empirical results on a collection of
highly non-projective datasets from Universal Dependencies show improvements
in accuracy over the projective approach of \citet{exact-minfeats}, as well as edge-factored maximum-spanning-tree
parsing. The results are on par with the 1-Endpoint-Crossing parser of \citet{pitler-1ec-3o}
(re-implemented under the same neural framework), but our algorithm is notably simpler and has
additional desirable properties: it is purely bottom-up, generalizable to higher coverage, and compatible
with transition-based semantics.

{%

\section*{Acknowledgments}
\label{sec:acknowledgments}

We thank the three anonymous reviewers for their helpful comments.
CG has received funding from the European
Research Council (ERC), under the European
Union's Horizon 2020 research and innovation
programme (FASTPARSE, grant agreement No
714150), from the TELEPARES-UDC project
(FFI2014-51978-C2-2-R) and the ANSWER-ASAP project (TIN2017-85160-C2-1-R) from MINECO, and from Xunta de Galicia (ED431B 2017/01).
TS and LL were supported in part by a Google Focused Research Grant to Cornell University.
LL was also supported in part by NSF grant SES-1741441. Any
opinions, findings, and conclusions or recommendations expressed in
this material are those of the author(s) and do not necessarily
reflect the views of the National Science Foundation or other
sponsors.
\par }

\bibliography{ref}
\bibliographystyle{acl_natbib}

\end{document}

\renewcommand{\thesection}{\Alph{section}}

\section{Implementation Details}
We use bi-LSTMs \cite{bilstm} for feature representations both at word and sentence level.
A $2$-layer bi-LSTM takes input from $64$-dimensional character embeddings,
and encodes intra-token information into its $128$ hidden units ($64$ for each direction).
Another $2$-layer bi-LSTM builds sentence-level context-sensitive features with the character LSTM encodings as inputs,
and assigns a $192$-dimensional vector representation to each word in the sentence.
All scoring functions for the edges/transitions are in the form of deep biaffine transformation \cite{dozat-iclr}.
For feature sets with more than two vectors, we define the score to be the sum of pairwise biaffine scores.
Scoring of $\{\bivec{g}, \bivec{h}, \bivec{m}, \bivec{si}\}$ in the baseline \oneec parser
is defined as the sum of biaffine scores of the follow pairwise interactions:
$\{\bivec{g}, \bivec{m}\}$, $\{\bivec{h}, \bivec{m}\}$, $\{\bivec{si}, \bivec{m}\}$.
Sum of biaffine scores for $\{\svec{1}, \svec{0}\}$ and $\{\svec{0}, \bvec{0}\}$ constitute the score for the three-vector feature set \sObZ.
All neural-network weight parameters are randomly initialized \cite{glorot2010init}, including the word and character embeddings.
We train each model
using Adam optimizer \cite{kingma-adam}
with initial learning rate $0.002$,
until the dev-set performance converges.
During training, dropout is applied to both multi-layer perceptrons in the deep biaffine functions and the recurrent connections \cite{dropout,gal-dropout}.
We set the keep rate to be $0.7$ everywhere.
Our implementation is based on the DyNet library \cite{dynet}.
Our code, including our re-implementation of the third-order \oneec parser
with neural scoring,
is available at \url{https://github.com/tzshi/mh4-parser-acl18}.

\bibliography{ref}
\bibliographystyle{acl_natbib}